\documentclass{article}

\usepackage{microtype}
\usepackage{graphicx}
\usepackage{mypackage}
\usepackage{booktabs} 

\usepackage{hyperref}

\usepackage[accepted]{icml2021}

\icmltitlerunning{Memory Efficient Neural Network Quantization via Implicit Differentiable-$k$-Means}

\begin{document}

\twocolumn[
\icmltitle{IDKM: Memory Efficient Neural Network Quantization via Implicit, Differentiable $k$-Means}



\icmlsetsymbol{equal}{*}

\begin{icmlauthorlist}
\icmlauthor{Sean Jaffe}{cs,ccdc}
\icmlauthor{Ambuj K. Singh}{cs}
\icmlauthor{Francesco Bullo}{ccdc}
\end{icmlauthorlist}

\icmlaffiliation{cs}{Department of Computer Science, UC Santa Bar
bara, Santa Barbara}
\icmlaffiliation{ccdc}{Center for
Control, Dynamical Systems, and Computation, UC Santa Bar
bara, Santa Barbara}

\icmlcorrespondingauthor{Sean Jaffe}{sjaffe@ucsb.edu}

\icmlkeywords{Quantization, Implicit Differentiation}

\vskip 0.3in
]



\printAffiliationsAndNotice{}  

\begin{abstract}
Compressing large neural networks with minimal performance loss is crucial to enabling their deployment on edge devices. \cite{cho2022dkm} proposed a weight quantization method that uses an attention-based clustering algorithm called differentiable $k$-means (DKM). Despite achieving state-of-the-art results, DKM's performance is constrained by its heavy memory dependency. We propose an implicit, differentiable $k$-means algorithm (IDKM), which eliminates the major memory restriction of DKM. Let $t$ be the number of $k$-means iterations, $m$ be the number of weight-vectors, and $b$ be the number of bits per cluster address. IDKM reduces the overall memory complexity of a single $k$-means layer from $\mathcal{O}(t \cdot m \cdot 2^b)$ to $\mathcal{O}( m \cdot 2^b)$. We also introduce a variant, IDKM with Jacobian-Free-Backpropagation (IDKM-JFB), for which the \textit{time complexity} of the gradient calculation is independent of $t$ as well. We provide a proof of concept of our methods by showing that, under the same settings, IDKM achieves comparable performance to DKM with less compute time and less memory. We also use IDKM and IDKM-JFB to quantize a large neural network, Resnet18, on hardware where DKM cannot train at all.
\end{abstract}

\section{Introduction}
 The capabilities of deep neural networks has grown in step with their size. The size of such neural networks is a major draw-back when used on edge-devices with limited memory and energy capacity. Building small neural networks that achieve high performance is thus an essential task for real-world applications. 
 
 Works such as MobileNet \cite{howard2017mobilenets} and SqueezeNet \cite{iandola2016squeezenet} aim to design explicitly compact neural structures. Efficient architectures can be found using neural architecture search, as in EfficentNet \cite{tan2019efficientnet}. Weight sharing such as repeated layers \cite{Dabre_Fujita_2019} methods also yield compact models. Implicit neural networks \cite{bai2019deep}, \cite{el2021implicit} take weight sharing to the extreme, implementing the behavior of deep neural networks with single recurrent layers. 

 Another means to generate compact neural networks is to compress large neural networks. Compression methods include pruning \cite{vadera2022methods}, knowledge-distillation \cite{hinton2015distilling}, and quantization. We focus on quantization, the aim of which is to limit a neural network to using a limited number of unique weights. The quantized weights are generally found with a clustering routine. 
 
 DKM \cite{cho2022dkm} achieved state-of-the-art results in quantization. The authors introduce a soft-$k$-means algorithm which uses an attention-based mechanism for assignment. DKM incorporates soft-$k$-means into the training loop and evaluates the loss of the model after clustering. Quantizing while training allows their algorithm to explicitly train weights which perform well after clustering. The soft assignments produce better gradients than hard assignments would. A major drawback is that DKM must store the result of every clustering iteration for the backward pass. This memory requirement is so restrictive, the clustering algorithm has to be stopped before convergence. 
 
 We solve the memory constraint of DKM by implementing the soft-$k$-means algorithm as a Deep Equilibrium Network (DEQ) \cite{bai2019deep}. DEQs use implicit differentiation to calculate the gradient for a layer independently of the forward pass. Implicit differentiation has been used in many novel recurrent architectures \cite{el2021implicit}, \cite{chen2018neural}. We employ implicit differentiation to calculate the gradient of the clustering algorithm in \textit{fixed memory} with respect to the number of iterations. Additionally, with Jacobian-Free Backpropagation \cite{fung2021jfb}, we can calculate the gradient in \textit{fixed time}. In total, our method can perform the $k$-means routine to greater precision while requiring less memory and time to calculate the gradient. As a result, we are able to perform extreme quantization on large models such as Resnet18 without expensive cloud infrastructure.

\section{Related Work}

\subsection{Compact neural Networks}
A number of works design neural networks to be explicitly compact, such as MobileNet \cite{howard2017mobilenets}, EfficientNet \cite{tan2019efficientnet}, and SqueezeNet
\cite{iandola2016squeezenet}, and ShuffleNet \cite{zhang2018shufflenet}. 

We are interested in compressing already-trained, high-performing neural networks.\cite{oneill2020compression} provides a thorough overview of neural network compression. Among the most well-studied neural network compression methods are pruning \cite{vadera2022methods}, knowledge-distillation \cite{hinton2015distilling} and weight sharing.

\subsection{Weight Sharing and Implicit Neural Networks}
\cite{Dabre_Fujita_2019} finds it beneficial to use the same weight in multiple layers. \cite{bai2019deep} take this to the extreme with their DEQ, which outputs the solution to a fixed point equation determined by a single neural network layer. Calculating the gradient of the DEQ requires implicit differentiation, which we discuss in section \ref{sec:gradient}.\cite{el2021implicit}  structure their DEQ as a state-space model.  \cite{winston2020monotone} and \cite{jafarpour2021robust} solve for the gradient of implicit, state-space models using monotone and non-euclidean monotone operator splitting methods, respectively. The Neural ODE \cite{chen2018neural} is another implicit model that uses a neural network to parameterize an ODE.

\subsection{Quantization of Neural Networks}

 An overview of quantization is provided in \cite{rokh2022comprehensive}. The literature on neural network quantization is vast. Quantization can roughly be split into two categories, post-training quantization (PQT), and  quantization aware training (QAT). PQT methods, such as \cite{banner2019post}, \cite{lin2017accurate}, and \cite{fang2020post}, quantize a pre-trained neural network without re-evaluation. QAT on the other hand, adjusts the model weights as it quantizes.  For example \cite{deepHan2015} uses pruning, quantization, and Huffman coding, with intermediary retraining steps. 

 Other QAT methods directly consider the gradient of the output of the quantized model to update the unquantized model. This is inherently difficult, as hard quantization is not differentiable. \cite{bengio2013estimating} introduce the widely used Straight-Through Estimator to approximate the gradients of hard non-linearities. Other works use cluster-based regularization \cite{wu2018deep}, \cite{ullrich2017soft}. Another work, \cite{stock2019and}, uses Product Quantization (PQ) to cluster subvectors of the weights.

 Neural networks can also be quantized heterogeneously layer-to-layer. \cite{wang2018hardware}  use reinforcement learning to identify hardware-specifc clustering schemes. \cite{chen2019metaquant} employs a meta-neural network to approximate the gradient of the non-quantized weights.\cite{park2020novel} uses an iterative training strategy which freezes layers subject to high instability. \cite{Dong_2019_ICCV} uses properties of the spectrum of the hessian to determine quantization schema.

\section{Neural Network Quantization}
 In what follows, we let $\norm{\cdot}$ denote the 2-norm. Suppose we have a deep neural network $f$ that takes input $x$ and outputs
a vector $y$ parameterized by weights $\textbf{W}\in \R^N$. For a given dataset, $\mathcal{D} = \{(x_i, y_i)\}$, the weight vector $w$ is trained to
minimize the loss:
\begin{equation}
    \label{eq:loss}
    \El(\textbf{W}) = \sum_{(x,y)\in \mathcal{D}}\norm{f(x,\textbf{W})-y}
\end{equation}

We now employ the Product Quantization setup, as described in \cite{stock2019and}. Consider a single layer with weights $W\in \R^n$ that is partitioned into $m=n/d$ sub-vectors $w_1,\dots,w_{m}$, each of which is an element of $\R^{d}$. The goal of neural network quantization is to learn a codebook of $k$, $d$-dimensional codewords $C = \{c_1,\dots, c_k\}$. Each element of $W$, $w_i$ is mapped to its closest codeword in $C$. Define the map $q:\R^d\times \R^{d\times k} \to\R^d$ such that $q(w_i, C)$ yields the closest codeword $c \in C$ to $w_i$. We also define $\mathbf{q}: \R^{d\times m}\times \R^{d\times k} \to\R^{d \times m}$ to be the map for which $\mathbf{q}(W, C)$ replaces every column $w_i$ of $W$ with $q(w_i, C)$.  Given a previously trained $W$, a codebook $C$ that minimizes the norm between each column in $W$ and their respective quantized weights can be found by minimizing the cost function:
\begin{align}\label{eq:clusterobj}
   \mathcal{C}(C, W) = \sum_{i=1}^m\norm{w_i - q(w_i, C)}^2
\end{align}

The $k$-means algorithm is guaranteed to converge to a local minimum of this objective. Hence, running $k$-means directly on previously trained weights is a known method for neural network quantization \cite{deepcompression}.

\subsection{Neural Network Quantization via Gradient Descent}
In model quantization, we are not just interested in minimizing the change in weights, but rather we want to minimize loss in performance as a result of quantization. To achieve this, we can incorporate the clustering loss \eqref{eq:clusterobj} into the original training loss as follows:
\begin{equation} \label{eq:combobj}
\begin{split}
 \\
 \El(W) = \sum_{(x, y) \in \mathcal{D}}\norm{f(x, \mathbf q(W, C)) - y} \\
 \text{with} \, C = \argmin_C \sum_i\norm{w_i - q(w_i, C)}^2
\end{split}
\end{equation}

The overall loss being minimized is the performance of the model using the quantized weights $\mathbf q(W, C)$ with optimal $C$. If $C$ is  found using $k$-means, a differentiable procedure, this objective can be minimized with stochastic gradient descent (SGD).

\subsection{Soft-$k$-means for Better Gradients} \label{sec:softkmeans}

While the $k$-means algorithm consists entirely of differentiable computations, the resulting gradient may lack useful information because of the hard assignment between weights and their respective centers. DKM uses continuous relaxation of the $k$-means algorithm. Rather than quantizing each $w_i$ to its closest codeword in $C$, the algorithm quantizes each $w_i$ by a convex combination of codewords. Let $r_\tau: \R^{d} \times \R ^{d \times k}  \to \R^{d}$ be the new quantization map, where $\tau>0$ is a temperature parameter. The formula for the quantizer $r_\tau$ is:
\begin{equation} \label{eq:softquant}
        r_\tau(w_i, C) = \sum_{j=1}^k a_{j}(w_i,C)c_j
\end{equation}
with
\begin{align}
    a_{j}(w_i,c)= \frac{\exp(-\norm{w_i - c_j}/\tau)}{\sum_{l=1}^k \exp(-\norm{w_i - c_l}/\tau)}\\ =  
  \big( \softmax_\tau(\big(-\norm{w_i - c_1},\dots,-\norm{w_i - c_k} \big)_j \label{eq:aterms}
\end{align}

The vector of weights $a(w_i, C)$ is the softmax of negative distances between $w_i$ and each codeword in $C$. Note that if $\tau=0$, then  $r_\tau(w_i, C) = q(w_i, C)$. As before, let $\mathbf{r}_\tau: \R^{d\times m}\times \R^{d\times k} \to\R^{d \times m}$ be the map for which $\mathbf{r}_\tau(W, C)$ replaces every column $w_i$ of $W$ with $r_\tau(w_i, C)$. Let $A: \R^{d \times m} \times \R^{d \times k} \to [0,1]^{m \times k}$ give the matrix with elements $A_{ij}(W,C) = a_j(w_i, C)$ as defined in equation \eqref{eq:aterms}. This distance-based attention matrix (introduced by \cite{bahdanau2014neural}) allows the matrix notation of $\mathbf{r}_\tau$:
\begin{equation} \label{eq:softquantmatrix}
    \mathbf{r}_\tau(W, C) = C\cdot A(W, C)^\top
\end{equation}

The codebook  $C$ can be computed with $EM$. The attention matrix $A(W,C)$ is computed in the \textit{E}-step. Distances between each weight and codeword are stored in a distance matrix $D(W, C)\in\Rnonnegative^{m \times k}$ defined by $d_{ij}(W, C) = \norm{w_i-
  c_j}$. We define \emph{row-wise softmax function}
$\rowsoftmax_\tau:\R^{m \times k} \to [0,1]^{m \times k}$ so that $A(W,C)$~\eqref{eq:softquant} is written in matrix form as:
\begin{equation}
\label{eq:arow}
  A(W,C) = \rowsoftmax_\tau(-D(W, C)).
\end{equation}

Centers are updated in the \textit{M}-step to be the average of all $w$, weighted by the center's importance to each $w$. Formally:
  \begin{equation}
    c_j^+ := \frac{\sum_{i=1}^m a_{j}(w_i,C)w_i}{\sum_{i=1}^m  a_{j}(w_i,C)}
  \end{equation}
or, in matrix notation:
\begin{equation}
\label{eq:centers}
    C^+ := \diag(A(W,C) ^\top\1_m)^{-1} A(W,C) ^\top W^\top
\end{equation}

The algorithm is described fully in algorithm \ref{alg:softkmeans}.
\begin{algorithm}[H]
  \caption{\textbf{Soft $k$-means }  
}
  \label{alg:softkmeans}
  \begin{algorithmic}[1]
    \REQUIRE{$W  \in \R^{d \times m}$ with columns $(w_1,\dots,w_m)$, tolerance parameter $\epsilon$ }
    \ENSURE{cluster centers $C$}
    
  \smallskip

  \STATE let $C\in\R^{k \times d}$ be an initial value of the cluster centers

  \REPEAT
  
  \STATE Compute the \emph{distance matrix} $D\in\Rnonnegative^{m \times
    k}$ by $d_{ij}:=\norm{w_i-c_j}$, $i\in\until{m}$ and $j\in\until{k}$

  \STATE Given a \emph{temperature} $\tau>0$, compute the \emph{attention
    matrix} $A$ by 
  \begin{equation*}
  A(W,C) = \rowsoftmax_\tau(-D(W, C))
  \end{equation*}
  \STATE Update the cluster centers by
  \begin{equation*}C^+ := \diag(A(W,C) ^\top\1_m)^{-1} A(W,C) ^\top W^\top
  \end{equation*}
  \UNTIL $\norm{C^+-C}<\epsilon$
  \STATE RETURN $C^+$
\end{algorithmic}
 \end{algorithm}
 \smallskip


We can redefine loss \eqref{eq:combobj} using the soft quantizer $r_\tau$ to be:
\begin{equation} \label{eq:softobj}
\begin{split}
 \\
 \El(W) = \sum_{(x, y) \in \mathcal{D}}\norm{f(x, \mathbf r_\tau(W, C)) - y} \\
 \text{with} \, C = \argmin_C \sum_i\norm{w_i - r_\tau(w_i, C)}^2
\end{split}
\end{equation}

\subsection{Memory Complexity of Differentiating Soft-$k$-means}
Suppose $C^*(W)$ is the optimal centers provided by soft-$k$-means \eqref{alg:softkmeans}. In order to optimize the objective \eqref{eq:combobj} via SGD, we will have to compute the gradient of soft-$k$-means, $\frac{\partial{C^*(W)}}{\partial W}$.  With backpropagation, this calculation requires saving every iteration of the algorithm in the autodiff computation graph and revisiting each iteration in the backward pass. Let $b = \lg(k)$ be the number of bits required to identify a cluster and $t$ be the number of iterations in soft-$k$-means. The GPU space necessary for a forward pass of a single soft-$k$-means layer is $\mathcal{O}(t\cdot m \cdot 2^{b})$. Thus, the memory required is dependent on the number of iterations $t$. Our contribution removes the need to store each iteration of soft-$k$-means in the forward pass for backpropagation, lowering the memory complexity to $\mathcal{O}( m \cdot 2^{b})$.

\section{Proposed Method}

Inspired by the work on deep equilibrium networks, as in \cite{bai2019deep}, we propose a method to calculate $\frac{\partial{C^*(W)}}{\partial W}$ efficiently. We use implicit differentiation to calculate the gradient of soft-$k$-means using only the centers found at convergence. The intermediate steps are not stored, so we can run the clustering algorithm for as many steps necessary without increasing the memory overhead. Additionally, by using Jacobian-Free Backpropagation(JFB) as in \cite{fung2021jfb}, we can calculate the backward pass with significantly reduced time complexity compared to the original backward computation and its corresponding DEQ implementation. In summation, our proposal is more efficient in time and space while providing more precise clustering than prior work. 

In section \ref{sec:fixedpoint} we derive a fixed point equation whose solution is that of soft-$k$-means. In section \ref{sec:gradient} we show how to take the gradient of that fixed point equation using only the solution. In section \ref{sec:JFB}, we show how to approximate that gradient with JFB. Our proposed method, IDKM, uses implicit soft-$k$-means to cluster weights inside of SGD. IDKM is described in full in algorithm \eqref{alg:train}.

\subsection{Defining the Fixed Point Equation} \label{sec:fixedpoint}
To motivate the implicit differentiation, we  will first frame the clustering algorithm \eqref{alg:softkmeans} as a fixed point equation. Thanks to \cite{bai2019deep}, we can derive the implicit computation of the gradient directly from the fixed point equation.
 Given a weight matrix $W\in\R^{m \times d}$ and a temperature $\tau>0$, with  $\1_m$ being the vector of ones in $\R^m$ and $rs_\tau(-D)$ being shorthand for $\rowsoftmax_\tau(-D(W,C))$ we define a map
$F: \R^{d\times k} \times\R^{d \times m}\to \R^{d\times k}$ :
\begin{equation}
  \label{eq:deepeqc}
  F \big(C, W) :=
  \diag(rs_\tau(-D)^\top\1_m)^{-1} rs_\tau(-D)^\top W^\top 
\end{equation}
The fixed point problem associated with equation \eqref{eq:deepeqc} is given by:
\begin{equation} \label{eq:fixedpoint}
    C^* = F(C^*, W)
\end{equation}
The solution $C^*$ is equivalent to the output of algorithm \ref{alg:softkmeans} by construction.

Note: While $F$ in equation \eqref{eq:deepeqc} is a function of $C$, one could also define a function which iterates on $A$ instead of $C$ that recovers the same solution $C^*$ (after one extra evaluation of equation \eqref{eq:centers}).

\subsection{Deriving the Gradient of the Fixed Point Equation}\label{sec:gradient}

We utilize Theorem 1 in \cite{bai2019deep} to derive the gradient of the solution to the fixed point equation \eqref{eq:fixedpoint}. For simplicity, we derive the gradient in the case when $d=1$.  Let $C^*(W)$ be the solution to the fixed point problem \eqref{eq:fixedpoint}. We first differentiate both sides of the equilibrium condition: $C^*=F(C^*, W)$:
\begin{align}
    &\frac{\partial C^*(W)}{\partial W} = \frac{\partial F(C^*(W), W)}{\partial W}\\
    &\frac{\partial C^*(W)}{\partial W} = \frac{\partial F(C^*, W)}{\partial W} + \frac{\partial F(C^*, W)}{\partial C^*}\frac{\partial C^*(W)}{\partial W}\\
&\Rightarrow \big(I_k - \frac{\partial F(C^*, W)}{\partial C^*}\Big)\frac{\partial C^*(W)}{\partial W} = \frac{\partial F(C^*, W)}{\partial W} \\
&\Rightarrow\frac{\partial C^*(W)}{\partial W} = \Big(I_k - \frac{\partial F(C^*, W)}{\partial C^*}\Big)^{-1} \frac{\partial F(C^*, W)}{\partial W} \label{eq:grad}
\end{align}

To solve\eqref{eq:grad}, we must compute the inverse matrix, which we denote by $M^*\in R^{k \times k}$:
\begin{equation} \label{eq:inv}
    M^* = \Big(I_k - \frac{\partial F(C^*, W)}{\partial C^*}\Big)^{-1} 
\end{equation}
Note that a sufficient condition for the existence of the inverse matrix above is that $ \frac{F(C^*, W)}{\partial C^*}$ has all eigenvalues with magnitude less than one. This is also the condition for the convergence of the fixed point equation \eqref{eq:fixedpoint}. Even the fixed point does not have global fixed points, we know there are local fixed points around which the matrix $ \frac{F(C^*, W)}{\partial C^*}$ behaves well. We can rearrange to get:\begin{align}
 M^* \Big(I_k - \frac{\partial F(C^*, W)}{\partial C^*}\Big)= I_k\\
    \Rightarrow M^* =  \frac{\partial F(C^*, W)}{\partial C^*}M^* + I_k \label{eq:backiter}
\end{align}

Let $G: \R^{k\times k} \to \R^{k\times k}$ be the map defined such that $G(M) =\frac{\partial F(C^*, W)}{\partial C^*})M + I_k$. It is clear from equation \eqref{eq:backiter} that $M^*$ is the solution to the fixed point equation: 
\begin{equation}\label{eq:backiter2}
    M^* = G(M^*)
\end{equation}

Using naive forward iteration, the iteration \eqref{eq:backiter2} convergence so long as all eigenvalues of $\frac{\partial F(C^*, W)}{\partial C^*}$ have magnitude less than one. This is also the condition for convergence of the forward iteration \eqref{eq:fixedpoint}.

We have shown so far how to calculate $\frac{\partial C^*(W)}{\partial W}$  independently of the computation of $C^*(W)$. At runtime, we take a solution $C^*(W)$, plug it into $F(C^*(W), W)$ one time and use autodiff to calculate $\frac{\partial F(C^*, W)}{\partial W}$. We then use some forward iteration solver to find the solution of the fixed point of equation \eqref{eq:backiter2}, which is equivalent to $\Big(I_k - \frac{\partial F(C^*, W)}{\partial C^*}\Big)^{-1}$.  Multiplying the two terms will yield $\frac{\partial C^*(W)}{\partial W}$. 

To Find $M$, we simply run the forward iteration above until convergence. For numerical convenience, we use an averaging iteration:
\begin{equation}
    M(t+1) = \alpha G(M(t)) + (1-\alpha) M(t)
\end{equation}
The smaller the $\alpha$, the more likely that the fixed point iteration will converge numerically, but the more iterations it will take to reach convergence. In our method, we set $\alpha=.25$ and if we see the iteration diverge, we start over and divide $\alpha$ by 2.

\subsection{Faster Gradient Computation with Jacobian-Free Backpropagation} \label{sec:JFB}

We further improve the efficiency of the gradient calculation by employing Jacobian-Free Backpropagation (JFB), as discussed in \cite{fung2021jfb}. JFB requires expanding the inverse term \eqref{eq:inv} using the Neumann series:
\begin{equation}
    M^* = \Big(I - \frac{\partial F(C^*, W)}{\partial C^*}\Big)^{-1} = \sum_{k=0}^{\infty}\Big(\frac{\partial F(C^*, W)}{\partial C^*}\Big)^k
\end{equation}
JFB uses the zeroth-order approximation of the Neumann series such that $M^* = I$. This approximation allows to entirely avoid solving the fixed point equation \eqref{eq:backiter2}. The gradient \eqref{eq:grad} now becomes: 
\begin{equation}\label{eq:jfbgrad}
\frac{dC^*(W)}{dW} = \frac{\partial F(C^*, W)}{\partial W} 
\end{equation}
 To motivate JFB further, the implicit function theorem, which enables us to calculate the gradient implicitly, ensures that the gradient of the solution to a fixed point equation is independent of the path taken to find the solution. In the case of the $k$-means, suppose a solution were found by initializing at an optimal $C^*$. Following algorithm \eqref{alg:softkmeans} from  $C^*$ would terminate after one iteration. The gradient of this solution path is exactly equation \eqref{eq:jfbgrad}. The principle of the gradient being independent of the solution path means this resulting gradient is the same as that calculated through any other solution path with any other initial $C$.

\begin{algorithm}[H]
  \caption{\textbf{IDKM }  
}
  \label{alg:train}
  \begin{algorithmic}[1]
    \REQUIRE{Dataset $\mathcal{D} = \{(x_i, y_i)\}$, pretrained neural network with weights $\mathbf{W}$, learning rate $\lambda$}
    \ENSURE{new weights $\mathbf{W}$}
    \WHILE{Still training}

    \FORALL{$(x, y)\in\mathcal{D}$}
    \STATE \emph{Turn off autodiff.}
    \FORALL{Layer $W\in \mathbf{W}$}
        \STATE Find $C^*(W)$ with soft-$k$ means \eqref{alg:softkmeans},
    \ENDFOR
    \STATE \emph{Turn on autodiff.}
    \STATE Calculate $\El(\mathbf{W})$ by \eqref{eq:softobj}.
    \FORALL{Layers $W\in \mathbf{W}$}
        \STATE Solve for $F$ as in \eqref{eq:deepeqc} one time, recover $\frac{\partial F(C^*(W), W)}{\partial W}$ .
        \STATE Compute $\Big(I - \frac{\partial F(C^*,W)}{\partial C^*}\Big)^{-1}$ by fixed point \eqref{eq:backiter2} or JFB \eqref{eq:jfbgrad} for IDKM-JFB. 
        \STATE Update $\frac{\partial C^*(W)}{\partial W}$ as in \eqref{eq:grad}
        \STATE Complete the calculation of $\nabla_{W}\El(W)$.
    \ENDFOR
    \STATE $\mathbf{W}^+ = \mathbf{W}- \lambda \nabla_\mathbf{W}\El(\mathbf{W})$
    \ENDFOR
    \ENDWHILE
  \STATE Return $\mathbf{W}^+$
\end{algorithmic}
 \end{algorithm}
 \smallskip

\section{Experiments}

To show a proof of concept of our method, we compare directly against DKM. We evaluate our method on two tasks in multiple compression regimes. In the following experiments, we use a learning rate of 1e-4 and temperature $\tau$, of 5e-4. We use an SGD optimizer with no momentum and train for 100 epochs. We let the clustering algorithm in all methods run until convergence or until 30 iterations, whichever comes first. 

\begin{table}[h]
\begin{tabular}{@{}llrrr@{}}
\toprule
k & d & \multicolumn{1}{l}{DKM} & \multicolumn{1}{l}{IDKM} & \multicolumn{1}{l}{IDKM-JFB} \\ \midrule
8 & 1 & 0.9615                 & \textbf{0.9717   }                   & 0.9702                          \\
4 & 1 &\textbf{ 0.9518 }                & 0.9501                         & 0.9503                          \\
2 & 1 & \textbf{0.7976   }               & 0.7701                       & 0.751                            \\
2 & 2 & 0.5512                  & \textbf{0.5822}                      & 0.5044                           \\
4 & 2 & \textbf{0.8688   }               & 0.825                        & 0.8444                           \\ \bottomrule
\end{tabular}
\caption{Top-1 accuracy of quantized two layer convolutional neural network pretrained on MNIST.}
\label{tab:Mnist}
\end{table}

\subsection{Small Convolutional Neural Network on MNIST}
First, we take a small, 2-layer convolutional neural network with 2158 parameters that has been pre-trained on MNIST \cite{lecun2010mnist} up to 98.4\% top-1 accuracy. We then run our implicit soft-$k$-means algorithm (IDKM), our algorithm with Jacobian-Free-Backpropagation (IDKM-JFB), and DKM\cite{cho2022dkm}. We vary the values of $d$ and $k$ to experiment with different compression ratios. The purpose of these experiments is to show our methods do not incur a performance drop by calculating the gradient implicitly. So, we let all methods run their clustering algorithm until convergence. We use such a small model because DKM would not be able to cluster until convergence on a large model, and our comparison would be unfair.

\begin{table}[h]
\begin{tabular}{@{}llrrr@{}}
\toprule
k & d & \multicolumn{1}{l}{DKM} & \multicolumn{1}{l}{IDKM} & \multicolumn{1}{l}{IDKM-JFP} \\ \midrule
8 & 1 & 3900                    & 2560                         & \textbf{1847}                    \\
4 & 1 & 1723                    & 1380                         & \textbf{1256}                    \\
2 & 1 & 1748                    & 1299                      & \textbf{1120}                    \\
2 & 2 & 1711                    & 1316                   & \textbf{1214}                            \\
4 & 2 & 1584                   & 1418                         & \textbf{1301}                    \\ \bottomrule
\end{tabular}
\caption{Time in seconds required to for each method to run for 100 iterations for varying compression schemes. Quantization target is a two-layer convolutional network pre-trained on MNIST.}
\label{tab:time}
\end{table}

Our results in Table~\eqref{tab:Mnist} show that IDKM performs fairly evenly to DKM. IDKM-JFB exhibits a slightly greater loss. This is expected because IDKM-JFB approximates the implicit gradient. We additionally show the total training time for each experiment in Table~\eqref{tab:time}. IDKM-JFB is the fastest method. Again, this result is expected because IDKM-JFB uses no iteration, implicit or otherwise, to calculate the gradient. Strikingly, IDKM also trains faster than DKM. This means that solving the fixed point equation \eqref{eq:backiter2} is faster than the autodiff's backpropagation through the soft-$k$-means iteration. We can claim from these results that, all else being equal, both of our methods are more  \textit{memory efficient} and \textit{faster} than DKM. IDKM-JFB allows for a further boost in speed with a small tradeoff in performance.

\subsection{Resnet18 on CIFAR10}

We experiment with clustering the 11,172,032-parmater Resnet18 \cite{he2016deep}. We first take a set of publicly-available weights, replace the last layer to yield 10 labels, and fine-tune on CIFAR10 \cite{krizhevsky2009learning} to a top-1 accuracy of 93.2\%. We then evaluate our quantization methods under the same compression regimes as before. Unlike with the 2-layer neural network, DKM will run out of memory for all values of $k$ and $d$ tested if more than $5$ iterations are used. In our experiments, $5$ clustering iterations or fewer will prevent the fully-quantized model from ever beating random. In this setting, it is not just significant that our methods perform well, but rather that they can perform at all. The results of IDKM and IDKM-JFB are shown in Table~\eqref{tab:resnet}.

\begin{table}[]
\centering
\begin{tabular}{@{}llrr@{}}
\toprule
k  & d & \multicolumn{1}{l}{\textbf{IDKM}} & \multicolumn{1}{l}{\textbf{IDKM-JFB}} \\ \midrule
2  & 1 & 0.5292                                & \textbf{0.5346}                           \\
4  & 1 & \textbf{0.897}                        & 0.8961                                    \\
8  & 1 & \textbf{0.9284}                      & 0.9273                              \\
2  & 2 & 0.3872                                & \textbf{0.4742}                           \\
4  & 2 & \textbf{0.897}                        & 0.8961                                    \\
16 & 4 & 0.8608          & \textbf{0.8648}                           \\ \bottomrule
\end{tabular}
\caption{Top-1 accuracy on Resnet18 model pretrained on CIFAR10 using our quantization methods. DKM never outperforms random assignment with the maximum iterations allowed by our hardware (5). So, its performance is not reported. Note that when $k=2$, the number of bits required for storage is $1$. With $k=2$ and $d=2$, half a bit is used for every weight in the quantized model.}
\label{tab:resnet}
\end{table}

\section{Discussion}

IDKM and IDKM-JFB remove the number of iterations as a factor in the memory complexity of Quantization methods that use soft-$k$-means during training. This reduction of memory is enough to justify our value of our methods to the research community. We additionally argue that allowing for more clustering iterations in the forward pass improves the overall performance of the algorithm. Due to hardware limitations, we have yet to compare directly against the presented results in \cite{cho2022dkm}. However, our experiments do allow us to extrapolate that our method will improve performance when used in a higher-resource setting. 

In settings where DKM method does not have the memory to run until convergence, \cite{cho2022dkm} simply limit the number of clustering iterations. The authors argue that this is not too restrictive on their performance, as the number of iterations  necessary to converge decreases in latter epochs as the model learns well-behaved weights. However, it is still desirable to allow for as many forward iterations as possible. We find a significant decrease in performance when the number of iterations are capped. For Resnet18, DKM was simply not able to learn quantizable weights with $5$ iterations, the same cap used by \cite{cho2022dkm}. It's a straight forward argument that DKM will perform better when allowed more clustering iterations.  IDKM exhibits comparable behavior to DKM under the same settings. So, just as in our Resnet18 experiments, we can expect that IDKM will outperform DKM when the former is allowed significantly more clustering iterations than the latter in general settings.

Enabling more clustering iterations allows us to consider more thoughtful tuning of the temperature parameter $\tau$.  \cite{cho2022dkm} are encouraged to lower the temperature to increase convergence speed. Of course, lowering the temperature lowers the quality of the clusters and of the gradient information. In the future, we would like to explore using higher temperatures equipped with annealing schemes to further improve performance.

IDKM and IDKM-JFB demonstrate extreme compression on a large-scale model using modest hardware. Additionally, our methods' memory efficiency widely expands the reasonable hyperparameter space available to train with soft-$k$-means, making it very likely we may find a configuration with a large improvement in performance in the large-model, large-dataset regime.

\textbf{Acknowledgements:} This work was supported in part by AFOSR grant FA9550-22-1-0059.



\bibliography{main}
\bibliographystyle{icml2021}





\end{document}